\documentclass{osa-article}

\journal{oe}

\articletype{Research Article}

\usepackage{microtype}                 
\PassOptionsToPackage{warn}{textcomp}  
\usepackage{textcomp}                  
\usepackage{times}                     
\usepackage{tabu}                      
\usepackage{booktabs}                  
\usepackage{amsmath}
\usepackage{amsfonts}
\usepackage{mathtools}
\usepackage{bbm}

\usepackage{comment}
\usepackage{siunitx}

\usepackage{bm}
\usepackage{multicol}
\usepackage{booktabs}
\usepackage{enumitem}

\usepackage{array}
\usepackage{longtable}
\usepackage{colortab}
\usepackage{colortbl}
\usepackage{arydshln}
\usepackage{xspace}
\newcommand{\etal}{et al.\@\xspace} 
\newcommand{\ie}{i.e.\@\xspace} 
\newcommand{\eg}{e.g.\@\xspace} 

\global\long\def\Matrix#1{\mathbf{\MakeUppercase{#1}}}

\global\long\def\Vector#1{\mathbf{\MakeLowercase{#1}}}

\global\long\def\Transpose{\mathrm{T}}

\global\long\def\Translation{\Vector t}

\global\long\def\CoordinateSymbol#1{#1}

\global\long\def\eyeCoordinate{\CoordinateSymbol E}

\global\long\def\displayCoordinate{\CoordinateSymbol D}

\global\long\def\Element#1#2{#1_{#2}}

\global\long\def\Real{\mathbb{R}}

\global\long\def\ImageCoords#1{\Element{\Vector u}{#1}}
\global\long\def\ImageCoordsAvg#1{\Element{\bar{\Vector u}}{#1}}
\global\long\def\ImageCoordsBase#1{\Element{\hat{\Vector u}}{#1}}
\global\long\def\EyePose{\Vector p}
\global\long\def\mapFunc{f}

\global\long\def\probability{\rho}
\global\long\def\ndfFunc{F_{\Theta}}
\global\long\def\encodeFunc#1{\gamma(#1)}

\usepackage{color}
\usepackage{savesym} \savesymbol{spacing} 
\usepackage{setspace}
\definecolor{Orange}{rgb}{1,0.5,0}

\definecolor{DarkGreen}{rgb}{0,0.5,0}
\definecolor{Purple}{rgb}{0.7,0,0.7}
\definecolor{Blue}{rgb}{0.2,0.2,0.8}
\definecolor{Red}{rgb}{1.0,0.0,0.0}
\definecolor{Brown}{rgb}{0.7,0.4,0.1}

\begin{document}
\title{Neural Distortion Fields for Spatial Calibration of Wide Field-of-View Near-Eye Displays}

\author{Yuichi Hiroi,\authormark{1, *} Kiyosato Someya,\authormark{2} and Yuta Itoh\authormark{1}}

\address{
\authormark{1} The University of Tokyo, 7-3-1, Hongo, Bunkyo-ku, Tokyo 113-8654, Japan \\
\authormark{2} Tokyo Institute of Technology, 4259 Nagatsutacho, Midoriku, Yokohama, Kanagawa 226-0026, Japan
}

\email{\authormark{*}yhiroi@g.ecc.u-tokyo.ac.jp} 



\begin{abstract}
We propose a spatial calibration method for wide Field-of-View (FoV) Near-Eye Displays (NEDs) with complex image distortions. Image distortions in NEDs can destroy the reality of the virtual object and cause sickness.
To achieve distortion-free images in NEDs, it is necessary to establish a pixel-by-pixel correspondence between the viewpoint and the displayed image. Designing compact and wide-FoV NEDs requires complex optical designs. In such designs, the displayed images are subject to gaze-contingent, non-linear geometric distortions, which explicit geometric models can be difficult to represent or computationally intensive to optimize.

To solve these problems, we propose Neural Distortion Field (NDF), a fully-connected deep neural network that implicitly represents display surfaces complexly distorted in spaces.
NDF takes spatial position and gaze direction as input and outputs the display pixel coordinate and its intensity as perceived in the input gaze direction.
We synthesize the distortion map from a novel viewpoint by querying points on the ray from the viewpoint and computing a weighted sum to project output display coordinates into an image. 
Experiments showed that NDF calibrates an augmented reality NED with 90$^{\circ}$ FoV with about 3.23 pixel (5.8 arcmin) median error using only 8 training viewpoints. Additionally, we confirmed that NDF calibrates more accurately than the non-linear polynomial fitting, especially around the center of the FoV.
\end{abstract}


\section{Introduction}\label{sec:intro}
\begin{figure}[tb]
 \centering
 \includegraphics[width=\columnwidth]{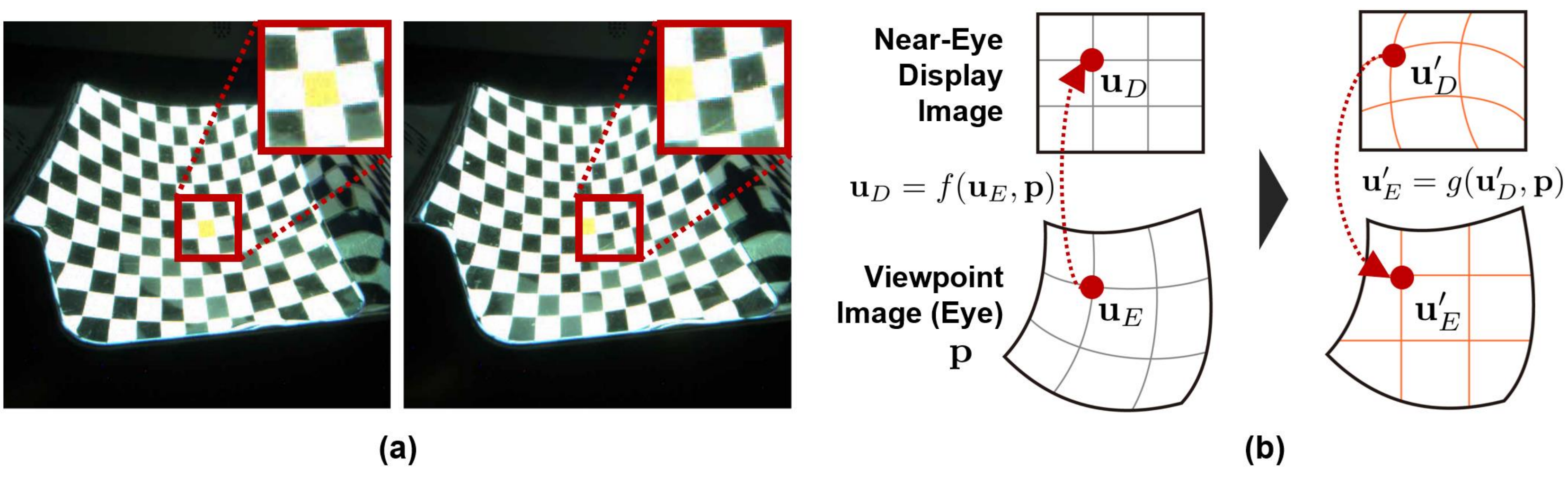}
 \caption{(a) Images of the checker pattern displayed on the wide-FoV AR-NED, captured from the viewpoint camera. Due to the complex beam combiner optics of the wide-FoV NED, the displayed image is subject to viewpoint-dependent image distortion. The images are captured from two different viewpoints with a distance of 6 mm along the x-axis. (b) Schematic diagram of the mapping between the display image and the retinal (or the viewpoint) image. (Left) When images are displayed directly on the NED, they appear distorted according to the user's viewpoint. The mapping function $\mapFunc$ represents the correspondence from the viewpoint coordinate to the display coordinate. (Right) If we correct the displayed image in advance, we can perceive the image without distortion.}
 \label{fig:viewpoint-shift}
\end{figure}

Near-eye displays (NEDs) provide an immersive visual experience in virtual reality (VR) and augmented reality (AR) by overlaying virtual images directly onto the user's view. 
Improving NED design to enhance its performance inevitably faces various trade-offs between optics design and image quality.
Achieving a wide field of view (FoV), a key requirement for NEDs to enhance immersion~\cite{koulieris2019near}, tends to compromise the form factor~\cite{hu2014high} or the image distortion to employ image expanding optics.
Modern NEDs employ a variety of off-axis optics for wide FoV such as curved beam splitter~\cite{akcsit2017near, dunn2017wide, guo2020raycast}, Holographic Optical Elements (HoE)~\cite{jang2017retinal, Kim2019}, and polarization-based pancake optics~\cite{maimone2020holographic, cakmakci2021pancake}.
While they improve the FoV and potentially reduce the display form factor, these complex, off-axis optics cause non-linear, viewpoint-dependent image distortion (Fig.~\ref{fig:viewpoint-shift}, a). Due to this distortion, a rectangle image on display appears distorted as a curved surface on the viewpoint.
Especially with wide-FoV NEDs, large image distortion occurs at the periphery of the display, causing the image to constantly sway as the eyes rotate and move, called pupil swim~\cite{geng2018pupilswim}. 
This image distortion and pupil swim can break the reality of virtual objects and, in the worst case, cause severe headaches and nausea.

This paper focuses on modeling this non-linear, viewpoint-dependent image distortion on wide-FoV NEDs. 
Let $\displayCoordinate$ and $\eyeCoordinate$ be the coordinate systems of the display and retinal image (or viewpoint camera image), respectively.
To formulate the image distortion, we want to know a map function $f$ that indicates which coordinates on the display image $\ImageCoords{\displayCoordinate}\in\Real^2$ appear on the coordinates in the retinal image $\ImageCoords{\eyeCoordinate}\in\Real^2$, as shown in Fig.~\ref{fig:viewpoint-shift} (b).
Especially for wide-FoV NEDs, this map varies not only with $\ImageCoords{\eyeCoordinate}$ but also with the translation and rotation of the eye. 
In this paper, we denote an eye pose by a 6D vector $\EyePose=[{\Vector v}, \Translation] \in \Real^6$ determined from the 3D rotation vector ${\Vector v} \in \Real^3$ and the position vector $\Translation \in \Real^3$.
Using these notations, a mapping function we focus on $\mapFunc:\Real^8\rightarrow\Real^2$ denotes as
\begin{eqnarray}
    \ImageCoords{\displayCoordinate} = f(\ImageCoords{\eyeCoordinate}, \EyePose).
    \label{eq:map-func}
\end{eqnarray}

\begin{figure}[tb]
 \centering
 \includegraphics[width=\columnwidth]{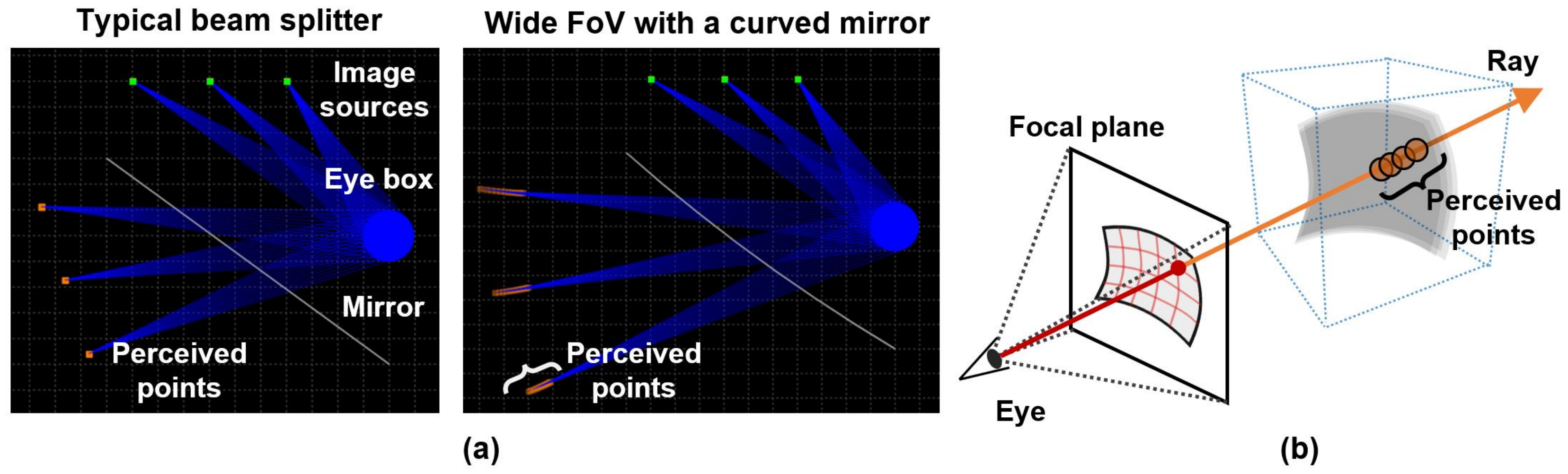}
 \caption{(a) A simplified optical simulation of NEDs employing different optics. (Left) Each display pixel is perceived as a point light source in a conventional flat beam splitter. (Right) When we use a curved mirror to expand the FoV, the perceived pixel gradually deviates from the point light source. For this simulation, we used a 2D ray optics simulator \cite{raytrace2022}. (b) Diagram of the relationship between the rays from the viewpoint and the perceived light source from the NED in the wide-FoV case in (a). In this case, we can model the perceived light sources as multiple translucent curved displays in space.}
 \label{fig:wide-fov-hmd}
\end{figure}

Theoretically, we can estimate $\mapFunc$ from the optical prescription of the NED. 
However, the deformation of the optical system and assembly is inevitable due to the manufacturing process and aging.
Thus, calibration of $\mapFunc$ is necessary for practical use.
In the case of NED with a typical beam splitter, the light emitted from each display pixel is perceived as a point light source at the viewpoint (Fig.~\ref{fig:wide-fov-hmd}, a). 
Hence, conventional work~\cite{klemm2017high} recovers $\mapFunc$ by estimating the 3D position of the point source of each display pixel using triangulation, then projecting it into a retinal image at a novel viewpoint.
However, in the case of wide-FoV NEDs, the virtual light source on each display pixel distribute in space due to the complex optics design of the projection system (Fig.~\ref{fig:wide-fov-hmd}, b). 
Therefore, accurate mapping with wide-FoV NEDs requires estimating the light field formed by the complex optics design.

Furthermore, in wide-FoV NEDs, image distortion changes dynamically with the viewpoint, making image distortion correction more challenging. Modeling and correcting static image distortion in VR-NED is a well-established technique~\cite{robinett1992computational, Rolland1993amethod}.
As a study of dynamic image distortion correction, 
the mainstream approaches extend the polynomial model for static image distortion correction to handle translations and rotations of the eye. 
Although some studies~\cite{hullin2012polynomial, schrade2016polynomials, martschinke2019gaze} have dealt with eye translation, the number of coefficients is insufficient to represent dynamic image distortion. Moreover, these studies do not take into account image distortions due to eye rotation, except for the method of approximating light ray field directly with a Gaussian polynomial fitting~\cite{itoh2015light}.

On the other hand, ray-tracing-based approaches~\cite{geng2018pupilswim, guo2020raycast} model image distortion by simulating the multi-stage refraction and reflection of the light rays passing through the optical system. Although these methods achieve high accuracy, they take a lot of computation time and are not practical for interactive VR/AR applications.
As a hybrid approach, concurrent with our work, Guan~\etal applied nonlinear dimensionality reduction to pre-traced light rays in a lens design application to simulate viewpoint-dependent image distortion in real time~\cite{guan2022perceptual}.

In contrast, we propose the \textit{implicit} representation model for the viewpoint-dependent image distortion, which is completely different from the previous approaches.
Our Neural Distortion Field (NDF) learns a distortion map $\mapFunc$ directly from a set of observed images without explicitly simulating the light field or optical aberration as polynomial models. 
NDF is an extension of the Neural Radiance Fields (NeRF)~\cite{mildenhall2020nerf}, a neural network-based representation model developed for novel view synthesis from multi-view color images. 
NeRF implicitly learns viewpoint-dependent light reflections and refractions and synthesizes novel-view images. 
Similarly, NDF is a neural network-based representation of the behavior of light rays passing from each pixel of a display through NED optics.
By using volumetric rendering for NDF representation, we can synthesize a non-linear, viewpoint-dependent  distortion map from a novel viewpoint.

The key novelty of our work is in applying NeRF's implicit, neural net-based view synthesis method to image distortion correction.
In the field of holography, Neural Holography~\cite{peng2020neuraholography, choi2021neural3d} implicitly represents optical misalignment, wave propagation, and display characteristics, and has significantly improved the quality of displayed images and processing time.
Similarly, our method aims to provide an another solution based on implicit function representation for the problem of image distortion modeling in NED, providing advantages such as improved processing time and accuracy.
Furthermore, it is expected that NDF can be incorporated into existing ray-tracing based image distortion correction and extended to a hybrid distortion representation model, similar to~\cite{guan2022perceptual}.
As a proof-of-concept and a preliminary step of the further model, in this paper we evaluate the accuracy of image distortion reproduction by a \textit{fully implicit} NDF model. 


\paragraph{Contributions.}
Our main contributions include the following:
\begin{itemize}
    \item We propose an NDF, the neural net model that can implicitly learn complex, view-point-dependent image distortion maps of NEDs directly from observed images.
    \item Experiments with using an off-the-shelf wide-FoV AR-NED show that NDF can simulate image distortion as accurately as or better than conventional non-linear polynomial mapping.
    \item We discuss improvements of NDF for applications in wide FoV HMDs and other optics designs and provide future research directions on image distortion correction with implicit representation models.
\end{itemize}


\section{Methods}\label{sec:method:NDF}

\begin{figure}[tb]
 \centering
 \includegraphics[width=\columnwidth]{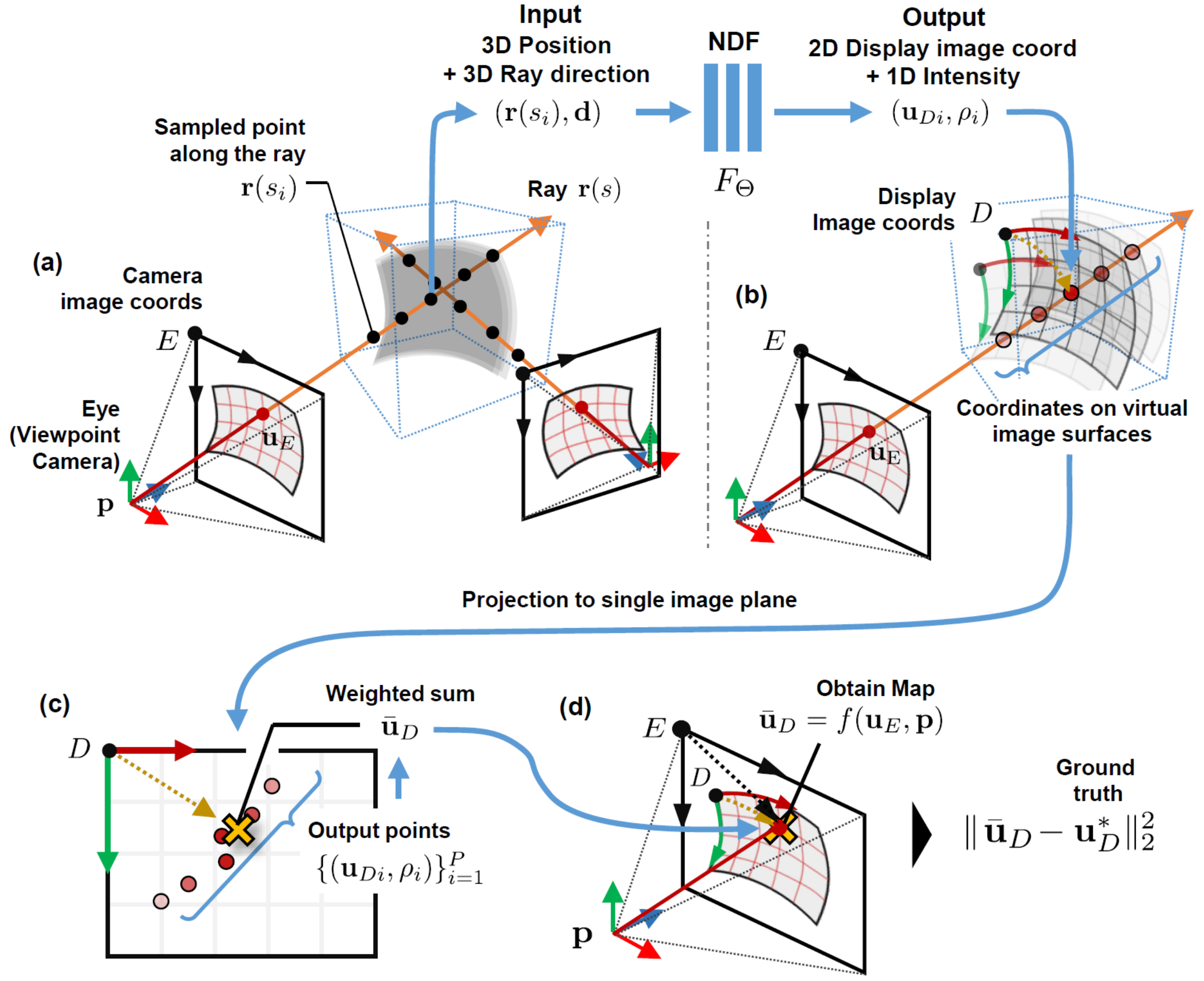}
 \caption{An overview of our NDF representation and distortion map generation. (a) We extend the ray connecting each pixel on the retinal image from the viewpoint and sample the points along the ray. NDF accepts as input the 3D coordinates of each point and ray direction. (b) The NDF returns the corresponding display coordinates and their intensity. The model assumes that there are myriad translucent curved displays in space, similar to Fig.~\ref{fig:wide-fov-hmd} (b). The output coordinates of the NDF represent which display pixels reach the eye at the input position and viewing direction. (c) We sum the output of the NDF at each point on the ray weighted by the intensity, and (d) estimate the subpixel-wise display coordinates perceived at the eye coordinate that consists of the ray. During training, we compute the loss between the estimating coordinate and the ground truth, and the loss is back propagated to NDF.}
 \label{fig:method-overview}
\end{figure}

In this section, we describe the basic NDF pipeline. Fig.~\ref{fig:method-overview} visualizes the overall pipeline of NDF. 
Sec.~\ref{sec:method-ndf} provides an overview of NDF representation. 
Sec.~\ref{sec:method-reconst-map} describes how to synthesize mapping $\mapFunc$ (Eq.~\ref{eq:map-func}) from outputs of NDF. 
Finally, Sec.~\ref{sec:method-nerf-tricks} describes how to train NDF from ground truths, including some tricks to improve NDF training.
Note that the description of NDF in this section follows the original NeRF.
With the recent development of NeRF research, various improved methods have already been proposed. Our implementation is based on the improved version of NeRF, which describes in Sec.~\ref{sec:impl-network}.


\subsection{Neural Distortion Field for Distortion Map Representation}\label{sec:method-ndf}
First, we describe our NDF representation. 
Briefly, when we look at a point from a certain direction, NDF returns the display pixel coordinate where the perceived light ray comes from and its intensity.
NDF is represented as a multi-layer perceptron (MLP) $\ndfFunc$, whose inputs are 5D coordinates (spatial position ${\Vector x} = [x, y, z]^\Transpose \in \Real^3$ and viewing direction $(\theta, \phi)$), and outputs are display coordinates $\ImageCoords{\displayCoordinate}$ and the intensity $\probability$ of the light source.
In practice, we express the viewing direction as a 3D Cartesian unit vector ${\Vector d} \in \Real^3$, \ie, $\ndfFunc:\Real^{6}\rightarrow\Real^3$.
Note that later in Sec.~\ref{sec:method-nerf-tricks}, we encode the input position ${\Vector x}$ into higher-order dimension $L$, \ie, $\ndfFunc:\Real^{L+3}\rightarrow\Real^3$.

We consider the ray connecting the eye position $\Translation$ and each pixel on the retinal image $\ImageCoords{\eyeCoordinate}$ (Fig.~\ref{fig:method-overview}, a). 
This ray denotes as ${\Vector r}(s) = \Translation + s{\Vector d}$. Since ${\Vector d} \in \Real^3$ is direction of the ray that moves in conjunction with eye rotation ${\Vector v}$, 
the ray ${\Vector r}(s)$ is essentially determined from $\ImageCoords{\eyeCoordinate}$ and $\EyePose=[{\Vector v}, \Translation]$.
When we sample a set of position and viewing direction $(\Vector r(s_i), \Vector d)$ on the ray at $s_i$ as an input of NDF, NDF $\ndfFunc$ outputs the display coordinate and the intensity $({\ImageCoords{\displayCoordinate}}_{i}, \probability_{i})$ (Fig.~\ref{fig:method-overview}, b).

Qualitatively, as the light from the microdisplay passes through the optical system, reflections and refractions create numerous transparent display surfaces in space, as shown in Fig.~\ref{fig:method-overview} (b).
From this, NDF can be regarded as implicitly learning these multiple, translucent display surfaces formed on the space.



\subsection{Distortion Map Reconstruction from Neural Distortion Field}\label{sec:method-reconst-map}
NDF outputs the display coordinates ${\ImageCoords{\displayCoordinate}}_{i}$ and its intensity $\probability_i$, which are the source of the light perceived at $({\Vector r}(s_i), {\Vector d})$.
By computing a weighted sum of the outputs along the ray ${\Vector r}(s)$, we can estimate the display pixel that is the source of the light perceived at each pixel on the retinal image $\ImageCoords{\eyeCoordinate}$ (Fig.~\ref{fig:method-overview}, c).
$\ImageCoordsAvg{\displayCoordinate}$ denotes this weighted sum of the display pixel coordinate.

Here, we sample $P$ points along the ray ${\Vector r}(s)$, which indexes as $\{s_i\}_{i=1}^{P}$ in order of proximity from the viewpoint.
NDF outputs $\{(\ImageCoords{\displayCoordinate i}, \probability_i )\}_{i=1}^{P}$ from these sampling points as input. 
Using the outputs, we calculate ${\ImageCoordsAvg{\displayCoordinate}}$ as
\begin{eqnarray}
\ImageCoordsAvg{\displayCoordinate}({\Vector r})=\sum^{N}_{i=1}\tau_i(1-\exp(-\probability_i\delta_i))\ImageCoords{\displayCoordinate i}, \quad \tau_i = \exp\left(-\sum^{i-1}_{j=1}\probability_j\delta_j\right),
\label{eq:volume-rendering}
\end{eqnarray}
where $\delta_i = s_{i+1}-s_{i}$ is the distance between adjacent samples. 
Note that from Sec.~\ref{sec:method-ndf}, since the ray ${\Vector r}$ is determined from $\ImageCoords{\eyeCoordinate}$ and $\EyePose$, Eq.~(\ref{eq:volume-rendering}) satisfies the form of Eq.~(\ref{eq:map-func}).


\subsection{Optimizing Neural Distortion Field}\label{sec:method-nerf-tricks}
By applying Eq.~(\ref{eq:volume-rendering}) to the entire field of view, the display coordinate system $\displayCoordinate$ is mapped as a 2D manifold on the retinal image $\eyeCoordinate$ (Fig.~\ref{fig:method-overview}, d).
To train NDF, we back-propagate the difference between the ground truth map obtained from several viewpoints and the map synthesized from Eq.~(\ref{eq:volume-rendering}).

Let $\ImageCoords{\displayCoordinate}^{*}({\Vector r})$ denote the ground truth of the display coordinates for each ray ${\Vector r}$.
From definition, the number of ${\Vector r}$ in single retinal image $\eyeCoordinate$ is equal to the number of pixels in $\eyeCoordinate$. 
In practice, we randomly sample a batch of ${\Vector r}$ from each pixel at each optimization iteration, then compute the total-squared loss $\mathcal{L}$:
\begin{eqnarray}
    \mathcal{L} = \sum_{{\Vector r}\in\mathcal{R}} \| \ImageCoordsAvg{\displayCoordinate}({\Vector r}) - \ImageCoords{\displayCoordinate}^{*} ({\Vector r}) \|^{2}_{2}
    \label{eq:positional-encoding}
\end{eqnarray}
where $\mathcal{R}$ denotes the set of randomly sampled rays.


The remainder of this subsection introduces improvements to more accurately simulate image distortion: positional encoding (Sec.~\ref{sec:positional-encoding}) and deviation map learning (Sec.~\ref{sec:deviation-map-earning}).

\subsubsection{Positional Encoding}\label{sec:positional-encoding}
Instead of training the NDF using Eq.~(\ref{eq:positional-encoding}), we introduce the technique called positional encoding, which is also used in the original NeRF, to facilitate the neural network to capture higher-order image distortions.
We encode the input position ${\Vector r}(s_i) \in \Real^3$ into higher-dimension vector $\encodeFunc{{\Vector r}(s_i)} \in \Real^L$.
Position encoding is generally represented by a combination of trigonometric functions~\cite{tancik2020fourfeat}, similar to the Fourier transform:
\begin{eqnarray}
{\Matrix P} = \begin{bmatrix}
1 & 0 & 0 & 2 & 0 & 0 & & 2^{L-1} & 0 & 0 \\
0 & 1 & 0 & 0 & 2 & 0 & \cdots & 0 & 2^{L-1} & 0 \\
0 & 0 & 1 & 0 & 0 & 2 & & 0 & 0 & 2^{L-1} \\
\end{bmatrix}^\Transpose, \quad \encodeFunc{\Vector x}=\begin{bmatrix} \sin(\Matrix P \Vector x) \\ \cos(\Matrix P \Vector x)
\end{bmatrix}.
\end{eqnarray}
By introducing the encoding, we redefine the MLP function as $\ndfFunc:\Real^{L+3}\rightarrow\Real^{3}$ and total-squared loss $\mathcal{L}$ as:
\begin{eqnarray}
    \mathcal{L} = \sum_{{\Vector r}\in\mathcal{R}} \| \ImageCoordsAvg{\displayCoordinate}(\encodeFunc{\Vector r}) - \ImageCoords{\displayCoordinate}^{*} (\encodeFunc{\Vector r}) \|^{2}_{2}.
\end{eqnarray}

Note that, compared to the original NeRF, which targets natural images, NDF deals with distorted image coordinates that vary relatively smoothly in space.
Thus, we are interested in the impact of encoding to high frequencies in NDF.
In later experiments (Sec.~\ref{sec:result-net-arch}), we evaluate the accuracy with different $L$ of the position encoding in NDF.


\subsubsection{Learning of Deviation Map from a Reference Viewpoint}\label{sec:deviation-map-earning}
In general, we normalize the raw data in the range [0, 1] to promote the training of neural networks.
In our NDF, the output of the neural network is the image coordinates.
For example, for a Full HD display, the range of raw output values in the horizontal direction is [0, 1920].
In this case, we should multiply the output value of the neural network by approximately $2.0\times10^3$.
This operation allows very small rounding errors ($< 0.001$) in the neural network to significantly affect the final results ($< 2$ pixel).
In the case of NeRF, even if the color changes slightly due to scaling, there is no significant perceptual difference. In the case of NDF, however, this difference appears as a perceptually significant distortion.

To avoid this, we set a reference eye pose $\hat{\EyePose}$ near the center of the eye box, and we use the measured display coordinates $\ImageCoordsBase{\displayCoordinate}$ at the reference viewpoint $\hat{\EyePose}$ as the reference map.
Then, we train the neural net $\ndfFunc$ using the deviation 
$\Delta\ImageCoords{\displayCoordinate}= \ImageCoords{\displayCoordinate}-\ImageCoordsBase{\displayCoordinate}$, instead of using the raw $\ImageCoords{\displayCoordinate}$.
In our training data set (Sec.~\ref{sec:acquire-dataset}), the range of $\Delta\ImageCoords{\displayCoordinate}$ is [-41.0, 39.5].
Thus, the scaling factor is 80.5, and we can reduce the effect of rounding errors in the neural network to 1/25 compared to the case using raw data.


\section{Implementation}\label{sec:impl-network}
To demonstrate our concept of implicit distortion map generation, we implemented NDF on top of mip-NeRF framework~\cite{barron2021mipnerf} implemented in JAX~\cite{jax2018github}.
mip-NeRF streamlines NeRF rendering by extending the NeRF query to be an expectation over a spatial region rather than a point, resulting in highly accurate and fast image reconstruction with reduced parameters.
Note that we currently select mip-NeRF based on the ease of implementation, training speed, and accuracy.
Hence, although performance can be improved by building on other NeRF frameworks, the underlying NDF concept (Sec.~\ref{sec:method:NDF}) remains unchanged.


\paragraph{Sampling Strategy on mip-NeRF.}
Instead of sampling each point on a ray, mip-NeRF samples a conical frustum connecting the viewpoint position and the pixel area.
As a result, mip-NeRF reduces unpleasant aliasing artifacts and improves the detail representation capability of NeRF.
With this improvement, the cone frustum around ${\Vector x}$ is considered a multivariate Gaussian distribution, and the mean value $\mathbbm{E}[\encodeFunc{\Vector x}]$ within the frustum is used as (integral) positional encoding.

\paragraph{Architecture.}
As an intensity network, we use an MLP with eight fully connected ReLU layers of 256 channels each. Then, we connect another MLP with four fully-connected ReLU layers of 128 channels each as the coordinate network in the latter stage.
The neural network architecture we chose utilizes the same configuration as NeRF~\cite{mildenhall2020nerf}, where this work is based. The original NeRF and derivative studies have adopted the same network architectures for controlled experiments, and our paper follows this convention. 
In NeRF, increasing the number of layers and channels from this number does not result in significant improvements in accuracy.
To accommodate NDF, we change the dimension of the output layer from the 3D color ${\Vector c}$ to the 2D coordinate $\Delta \ImageCoords{\displayCoordinate}$ from the mip-NeRF code base.

In the original NeRF, to reduce the influence of the view direction on the output intensity $\probability$, in the connection part of the network, $\mathbbm{E}[{\Vector d}]$ is input later after we extract $\probability$ from the network at the former stage.
We also adopt the same two-stage architecture for NDF, because we consider that the directivity of light emitted from a display does not change significantly with minute changes in angle.

In mip-NeRF, the activation functions used to generate the color ${\Vector c}$ (in NDF, map $\Delta \ImageCoords{\displayCoordinate}$) and intensity $\probability$ were sigmoid and softplus, respectively.
There are possible candidates for the activation function. 
As the activation function for color (in NeRF) or coordinate (in NDF) output, mip-NeRF used sigmoid to suppress the output value ${\Vector c}$ to the [0, 1] floating-point RGB color space. 
Instead, we consider that the piecewise-linear function such as ReLU is appropriate for the activation function, because NDF outputs the coordinate value $\Delta \ImageCoords{\displayCoordinate}$. 
Also, as the activation function for intensity output, the original NeRF uses SoftPlus. 
Instead, we consider Sigmoid as the possible candidate because we consider it would be better to adopt a stochastic model, considering that the light emitted from each point of the display is gradually dispersed from 100\%.
Based on these hypotheses, we evaluate the impact on accuracy when learning with different activation functions in Sec.~\ref{sec:result-net-arch}.

\paragraph{Training.}
We train NDF by Adam~\cite{kingma2015adam} with a batch size of 1024 and a learning rate $\{\eta_{i}\}$ that is annealed logarithmically from $\eta_{0}=5 \cdot 10^{-4}$ to $\eta_{n}=5 \cdot 10^{-6}$. 
We train all NDF models up to $5.0 \times 10^5$ iterations, where the training error is no smaller on a logarithmic scale. Later in Sec.~\ref{sec:result-net-arch}, we evaluate the relationship between the number of training iterations and accuracy in detail.

Our network takes about 4 hours to train on an NVIDIA RTX 3090 GPU and about 20 seconds to generate the whole distortion map. We can accelerate the map generation in real time with the latest NeRF architecture, as later discussed in Sec.~\ref{sec:discussion}.

\section{Data Acquisition}\label{sec:acquire-dataset}
To compare NDF with other mapping estimation methods, we acquire a data using a commercial wide FoV AR-NED.
Sec.~\ref{sec:data-hardware} describes the hardware setup for capturing NEDs from different viewpoints. 
Then, Sec.~\ref{sec:data-viewpoint-camera} describes the viewpoint camera locations and sampling intervals for training and testing. Finally, Sec.~\ref{sec:data-obtain-map} describes how to obtain the correspondence between viewpoint image coordinates and display coordinates at each viewpoint.

\subsection{Hardware Setup}\label{sec:data-hardware}
\begin{figure}[tb]
 \centering
 \includegraphics[width=0.6\linewidth]{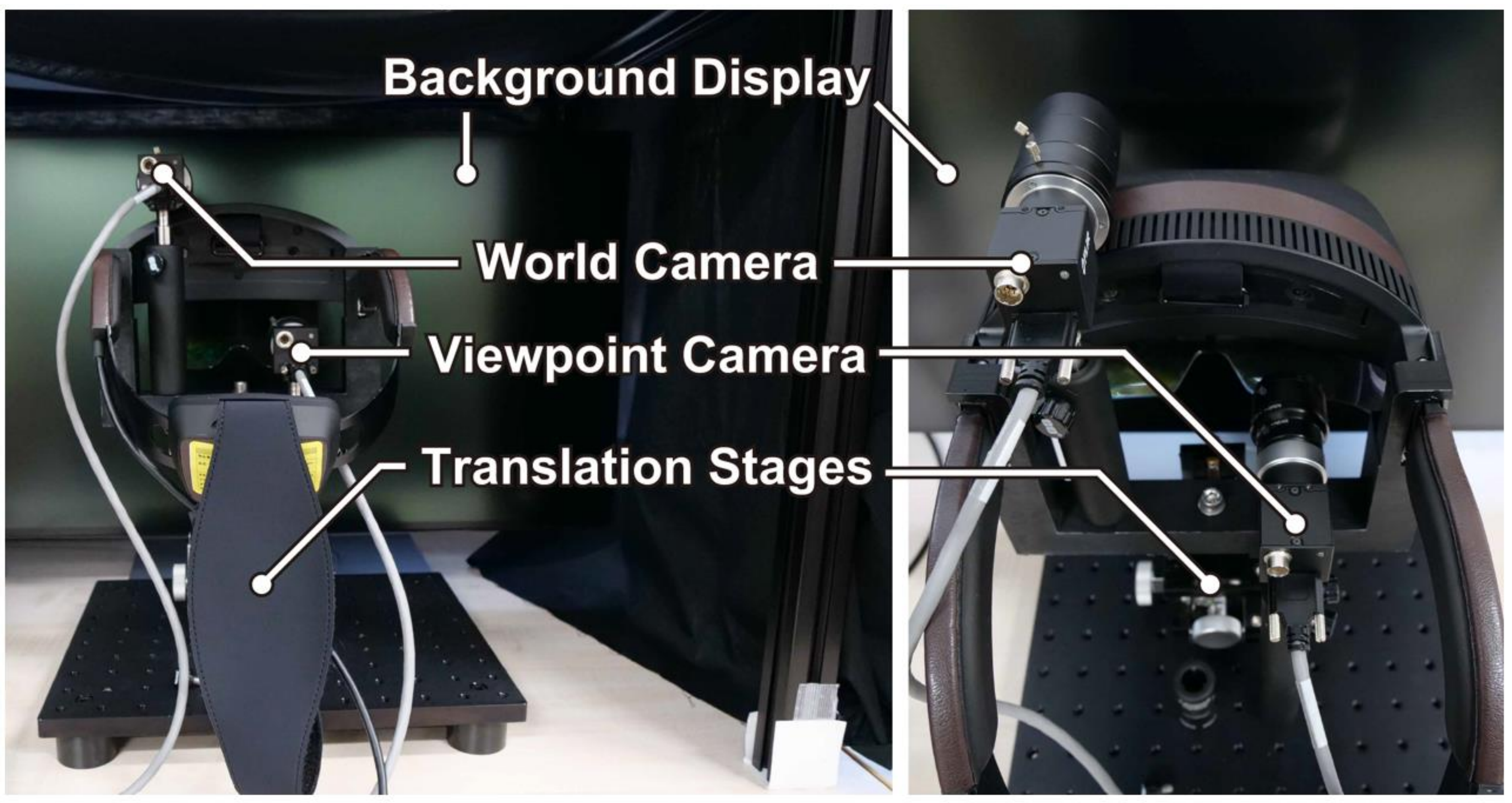}
 \caption{The hardware setup for the calibration. (Left) A back view of the calibration setup. (Right) Bird's eye view of the setup. The OST-HMD and two cameras are rigidly mounted each other and placed on an XYZ stage. The display is placed in front of the system and used as calibration reference. }
 \label{fig:hardware-overview}
\end{figure}

Figure~\ref{fig:hardware-overview} (a) shows the hardware setup of our experiment. We use a Meta 2 (Meta Company, 90$^\circ$ FoV) as a wide-FoV AR-NED with curved beam combiners, a Dell U2718Q as a background display, and two Blackfly S Color 12.3 MP USB3 cameras as a viewpoint camera and a world camera, respectively.
We mount the AR-NED and the world camera on a composite translation stage, which moves in x-, y-, and z-direction, respectively.  
We fix the position of the viewpoint camera and the background display with printed 3D jigs to move the OST-HMD with respect to them. In other words, the viewpoint camera position is translated relative to the NED, and the viewpoint camera position is translated relative to the NED, and the world camera position is treated as the origin.
To prevent the background display reflects the light from room lights, we cover the entire setup with black cloth.

\subsection{Viewpoint Camera Positions}\label{sec:data-viewpoint-camera}
\begin{figure}[tb]
 \centering
 \includegraphics[width=\linewidth]{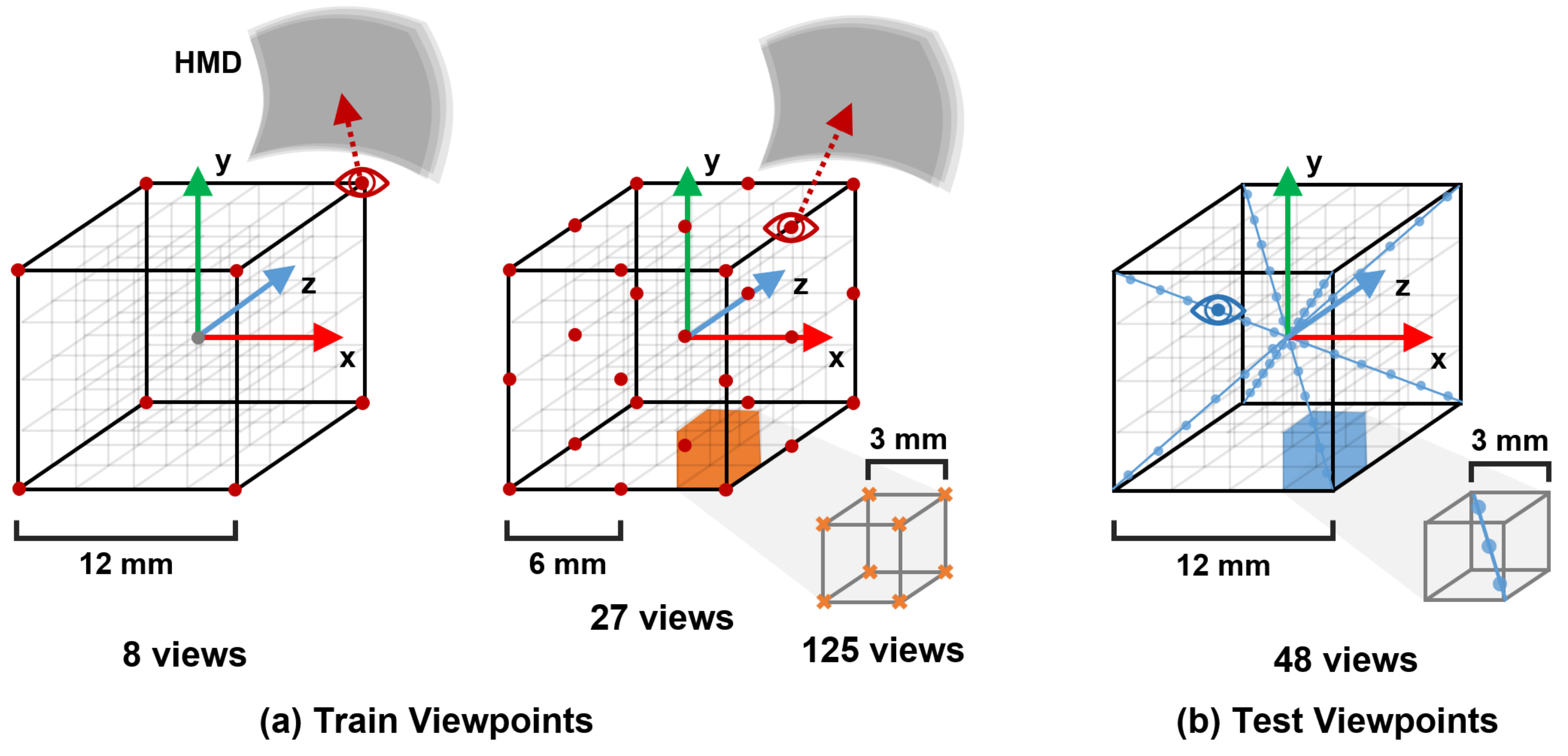}
 \caption{Viewpoint camera positions for data acquisition. (a) Camera positions for training data. For the full training set (125 views), we place the viewpoint camera on each vertex of a 3 mm square lattice. Then, to create training data for a fewer number of viewpoints, we extract training data for 8 and 27 viewpoints on the vertices indicated by red dots. (b) Camera positions for test data. We sample 12 points from the diagonal of a 12 mm square eyebox indicated by blue dots, a total of 48 viewpoint positions, as the test dataset. To avoid positional overlap with the training data set, we divide the diagonal of the 3 mm square cube into six equal parts and extract the 1st, 3rd, and 5th points.}
 \label{fig:dataset-overview}
\end{figure}
Before obtaining the coordinate transformation map between the HMD and the viewpoint camera, we set the measuring viewpoint positions for both training and testing.

Fig.~\ref{fig:dataset-overview} shows the viewpoint camera positions.
For training, we get the data from 125 viewpoints at the are vertices of grid cubes inside an eyebox cube divided into $4^3$ cubes, as shown in Figure~\ref{fig:dataset-overview} (a). 
Each grid cube is 3 mm on one side, so the entire eyebox cube is 12mm on one side. We use these $5^3 = 125$ datasets as training data.
In the experiment, we also evaluated the accuracy of each method on a dataset with a wider interval (\ie, fewer viewpoints). With a gap of 6 mm, the number of study viewpoints is $3^3 = 27$. Also, with an interval of 12 mm, we use only the $2^3 = 8$ viewpoints that make up the eyebox as training data.

We get the data from 48 viewpoints for testing, as shown in Fig.~\ref{fig:dataset-overview} (b).  
We sample 12 points from each diagonal of a 12 mm square cube representing the entire eye box and use these 48 points as viewpoint positions for testing.
This sampling interval for testing is based on~\cite{martschinke2019gaze}, which is designed so that the distribution of test points covers the entire eyebox and is as uniform as possible.

To detect the poses of the viewpoint camera, we display a 38.85 mm AR marker of a $4 \times 4$ binary pattern on the background display.
Then, we obtain the viewpoint camera pose $\EyePose$ as a relative posture with the world camera as the origin.
Since we manually adjust the translating stage position this time, there is a slight discrepancy between the ideal viewpoint position described above and the actual measurement position.
However, this slight discrepancy from the ideal does not affect the training process because we train the MLP with the viewpoint positions obtained from the actual measurements. 

\subsection{Obtaining Map at Each Viewpoint}\label{sec:data-obtain-map}
On each viewpoint, we obtain the correspondence between the coordinate system on the viewpoint camera and the display coordinate system.
Let $N$ be the number of training viewpoints and $\{ \EyePose^{*}_i = [{\Vector v}^{*}_i, {\Translation}^{*}_i]\}_{i=1}^{N}$ be the set of eye pose at the training viewpoints.
At each training viewpoint $\EyePose^{*}_i$, we obtain the ground truth of the mapping function from $\ImageCoords{\eyeCoordinate}$ to $\ImageCoords{\displayCoordinate}$, which denotes $f^{*}_i:\Real^2\rightarrow\Real^2$.
This $f^{*}_i$ can also be regarded as $\mapFunc$ given the eye posture $\EyePose^{*}_i$, \ie,
\begin{eqnarray}
\label{eq:mapfunc-trainings}
\ImageCoords{\displayCoordinate}=f^{*}_i(\ImageCoords{\eyeCoordinate})=\mapFunc(\ImageCoords{\eyeCoordinate}, \EyePose^{*}_i).
\end{eqnarray}

To establish a set of the map at the training viewpoints $\{f_i^{*}\}_{i=1}^{N}$, 
we display the gray-code pattern images and capture them on the viewpoint camera at first.
Then, we obtain the discrete correspondence between viewpoint coordinates and display coordinates from the gray code image as a look-up table (LUT).
At viewpoint $\EyePose^{*}_i$, $J_i$ denotes the number of pairs for which we can obtain a correspondence between the coordinates  and $\{(\ImageCoords{\eyeCoordinate ij}, \ImageCoords{\displayCoordinate ij})\}_{j=1}^{J_i}$ denotes the LUT of coordinate pairs.

Then we apply Gaussian kernel regression to interpolate this LUT as a continuous polynomial function $\mapFunc^{*}_{i}$ \cite{itoh2015light}. We express $\mapFunc^{*}_{i}$ as a Gaussian polynomial model:
\begin{eqnarray}
\mapFunc^{*}_{i}(\ImageCoords{\eyeCoordinate})= {\Matrix A^\Transpose}{\Vector \phi}(\ImageCoords{\eyeCoordinate}) = \sum^{K}_{k=1}
\boldsymbol{\alpha}_{k}^\Transpose \phi_k(\ImageCoords{\eyeCoordinate}),\quad 
\phi_k(\ImageCoords{\eyeCoordinate}) = \exp\left(\frac{-(\ImageCoords{\eyeCoordinate}-\boldsymbol{\mu}_{k})^\Transpose(\ImageCoords{\eyeCoordinate}-\boldsymbol{\mu}_{k})}{2\sigma^2}\right)
\label{eq:gaussian-interpolation}
\end{eqnarray}
where $\boldsymbol{\phi}$ is Gaussian radial basis vector, $K$ is the number of basis functions, $\sigma$ is the kernel width, $\{\boldsymbol{\mu}_k\}$ is the center of Gaussian kernel (randomly chosen from $\{\ImageCoords{\eyeCoordinate ij}\}$), and ${\Matrix A} = [\boldsymbol{\alpha}_{1}, \cdots, \boldsymbol{\alpha}_{k}, \cdots, \boldsymbol{\alpha}_{K}]^\Transpose$ is a $K \times J_i$ coefficient matrix. 
After that, we determine ${\Matrix A}$ using the regularized least-square estimator:
\begin{eqnarray}
{\Matrix A} = (\boldsymbol{\Phi}^\Transpose \boldsymbol{\Phi} + \lambda {\Matrix I}_{K} )^{-1}\boldsymbol{\Phi}^\Transpose{\Matrix U}_{\displayCoordinate}
\label{eq:train-map-regression}
\end{eqnarray}
where $\boldsymbol{\Phi}$ is a $J_i \times K$ design matrix defined as $[\boldsymbol{\Phi}]_{jk}={\Vector \phi}_k(\ImageCoords{\eyeCoordinate ij})$, $\lambda$ is the regularization parameter, ${\Matrix I}_{K}$ is a $K \times K$ identity matrix, and ${\Matrix U}_{\displayCoordinate} = [\ImageCoords{\displayCoordinate i1}, \cdots,  \ImageCoords{\displayCoordinate ij}, \cdots, \ImageCoords{\displayCoordinate iJ_i}]^\Transpose$.
We implement Eq.~(\ref{eq:train-map-regression}) using MATLAB R2022a.
We repeat the above operations for all training viewpoints $\{\EyePose^{*}_{i}\}_{i=1}^{N}$ to obtain a set of ground truth maps $\{\mapFunc_i^{*}\}_{i=1}^{N}$. Additionally, we obtain the ground truth maps for all test viewpoints for evaluation.

\section{Experiments}
Using the acquired dataset, we compared the accuracy of NDF with other map interpolation methods (Sec.~\ref{sec:experiments-methods}) and evaluate the performance of NDF. 
We applied each method to the dataset and generated a map at the test viewpoints.
Then, we evaluated quantitatively with respect to the reprojection error with respect to the ground-truth (Sec.~\ref{sec:experiment-between-methods}).
After that, we evaluated the reproducibility of dynamic image distortion with respect to the viewpoint image in the FoV (Sec.~\ref{sec:experiment-difference-fov}) and the spatial distribution of the test viewpoints (Sec.~\ref{sec:experiments-difference-viewpoints}), respectively.
Finally, we evaluated the difference in accuracy when changing the network configuration of NDF (Sec.~\ref{sec:result-net-arch}).

\subsection{Interpolation Methods for Comparison}\label{sec:experiments-methods}
First, prior to the evaluation, we briefly discuss each of the other interpolation methods being compared.
The problem set in this paper is to interpolate a map $\mapFunc:\Real^8\rightarrow\Real^2$ at a novel viewpoint $\EyePose$ from the ground truth maps at the training viewpoints $\{f^{*}_i\}_{i=1}^{N}$ (Eq.~(\ref{eq:mapfunc-trainings})).
We implemented 3 interpolation methods in addition to NDF: (i) 3D reconstruction, (ii) linear interpolation and (iii) Gaussian (non-linear) polynomial interpolation.
Note that, except for the 3D reconstruction-based interpolation, we ignore the eye rotation ${\Vector v}$. In other words, we train $\hat{\mapFunc}(\ImageCoords{\eyeCoordinate}, \Translation):\Real^5\rightarrow\Real^2$, instead of the complete $\mapFunc$.

\paragraph{(i) 3D Reconstruction of Virtual Display Surface.}
Assuming that each pixel $\ImageCoords{\displayCoordinate}$ on the display image forms a virtual display surface in space, we recover the 3D point of each pixel by triangulation and bundle adjustment~\cite{klemm2017high}. 
Then we estimate the map $\mapFunc$ by re-projecting the reconstructed 3D surface onto the image plane $\ImageCoords{\eyeCoordinate}$ at the new viewpoint $\EyePose$.
As discussed in Sec.~\ref{sec:intro}, this model assumes each pixel as a point light source.

\paragraph{(ii) Linear Interpolation.}
We take the 8 viewpoints on the cubic grid containing the new viewpoint position $\Translation$ from the training data (Fig.~\ref{fig:hardware-overview}, b).
Then, we estimate $\hat{\mapFunc}(\ImageCoords{\eyeCoordinate}, \Translation)$ using tri-linear interpolation using $\mapFunc^*_i(\ImageCoords{\eyeCoordinate})$ at the 8 vertices of the cubic grids.

\paragraph{(iii) Non-Linear Gaussian Polynomial Fitting.}
We learn $\hat{\mapFunc}(\ImageCoords{\eyeCoordinate}, \Translation):\Real^5\rightarrow\Real^2$ directly from the Gaussian polynomial model (Eq.~(\ref{eq:gaussian-interpolation})), using the ground truth maps $\{f^{*}_{i}(\ImageCoords{\eyeCoordinate}):\Real^2\rightarrow\Real^2\}_{i=1}^{N}$ and corresponding eye positions $\{{\Vector t}^{*}_i\in\Real^3\}_{i=1}^{N}$.

\subsection{Reprojection Error between Distortion Models}\label{sec:experiment-between-methods}
We trained maps using each method with $N=8, 27, 125$ training viewpoints, and we calculated the reprojection error of the output display coordinates of each pixel at 48 test viewpoints. 
We used only pixels within the area where the AR-NED can be seen on each viewpoint image for error calculation. 
In addition to the per-pixel error, we calculated the angular error from the viewpoint at each pixel.
NDF in this subsection were trained with the position encoding $L=16$ and the output coordinate and intensity activation functions ReLU and SoftPlus, respectively.

\begin{figure}[tb]
 \centering
 \includegraphics[width=\linewidth]{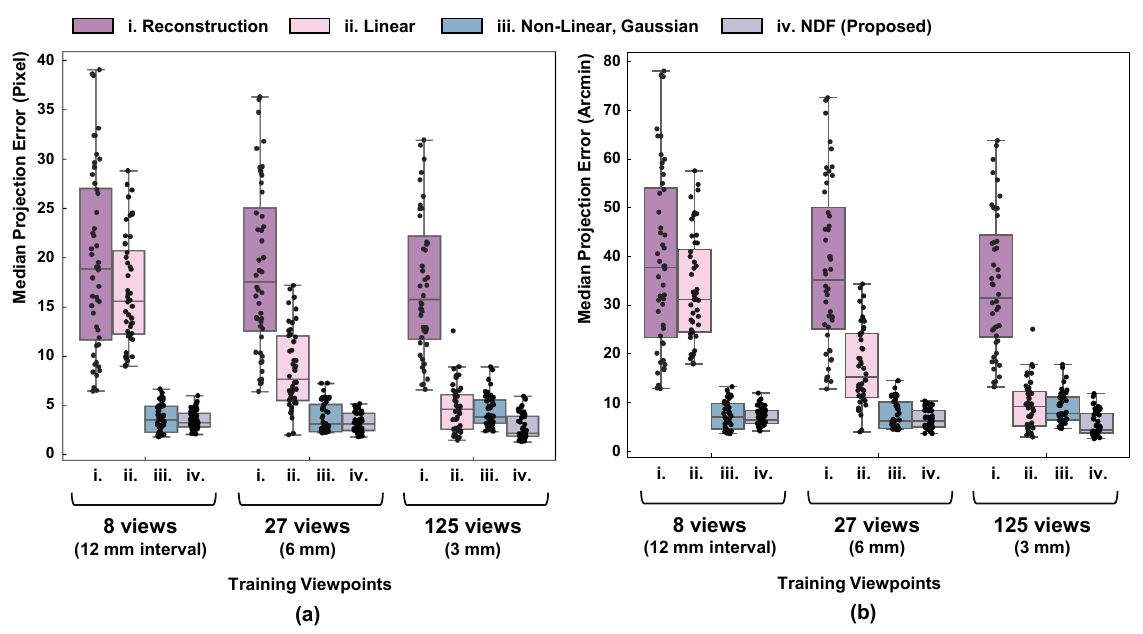}
 \caption{The re-projection error of $\ImageCoords{\displayCoordinate}$ on for different training viewpoints and interpolation methods. We plotted the median error at each viewpoint for 48 test viewpoints. (a) Summary of median projection errors in pixels. (b) Summary of median angular errors.}
 \label{fig:result-compare-methods}
\end{figure}

Figure~\ref{fig:result-compare-methods} shows the pixel errors and median angular error for different training viewpoints and interpolation methods. In $N=8$, the mean errors are (i) 18.86 pixel (31.96 arcmin), (ii) 15.60 pixel (24.86 arcmin), (iii) 3.64 pixel (5.99 arcmin), (iv) 3.23 pixel (5.79 arcmin). From the results, we confirm that NDF can recover maps with accuracy comparable to non-linear polynomial-based fitting with very few training viewpoints.
Also, the deteriorating accuracy of the (i) 3D reconstruction-based method confirms that the spatially distributed model of light sources assumed by the (iv) NDF correctly approximates the optical model of the wide-FoV NED.

From Fig.~\ref{fig:result-compare-methods}, while nonlinear polynomial-based methods do not change significantly as the number of training viewpoints increases, NDF shows an improvement in accuracy as the number of training viewpoints increases: (iii) 2.63 pixel (5.38 arcmin) vs. (iv) 3.13 pixel (5.42 arcmin) in $N=27$, and (iii) 3.61 pixel (6.46 arcmin) vs. (iv) 2.17 pixel (4.25 arcmin) in $N=125$. 
Moreover, in all cases of $N=8, 27, 125$,  (iii) Gaussian polynomial fitting has a larger error variance in the test viewpoints than (iv) NDF.
From the result, we assume that in the case of polynomial-based explicit optimization, the more points are trained, the more they overfit the map at the center of the eyebox. In contrast, in NDF, the implicit function representation uniformly performs the optimization. 
Later we quantitatively analyse this difference in error distribution between test viewpoint positions in Sec.~\ref{sec:experiments-difference-viewpoints}.

\subsection{Error Distribution in the Field of View}\label{sec:experiment-difference-fov}
\begin{figure}[tb]
 \centering
 \includegraphics[width=\linewidth]{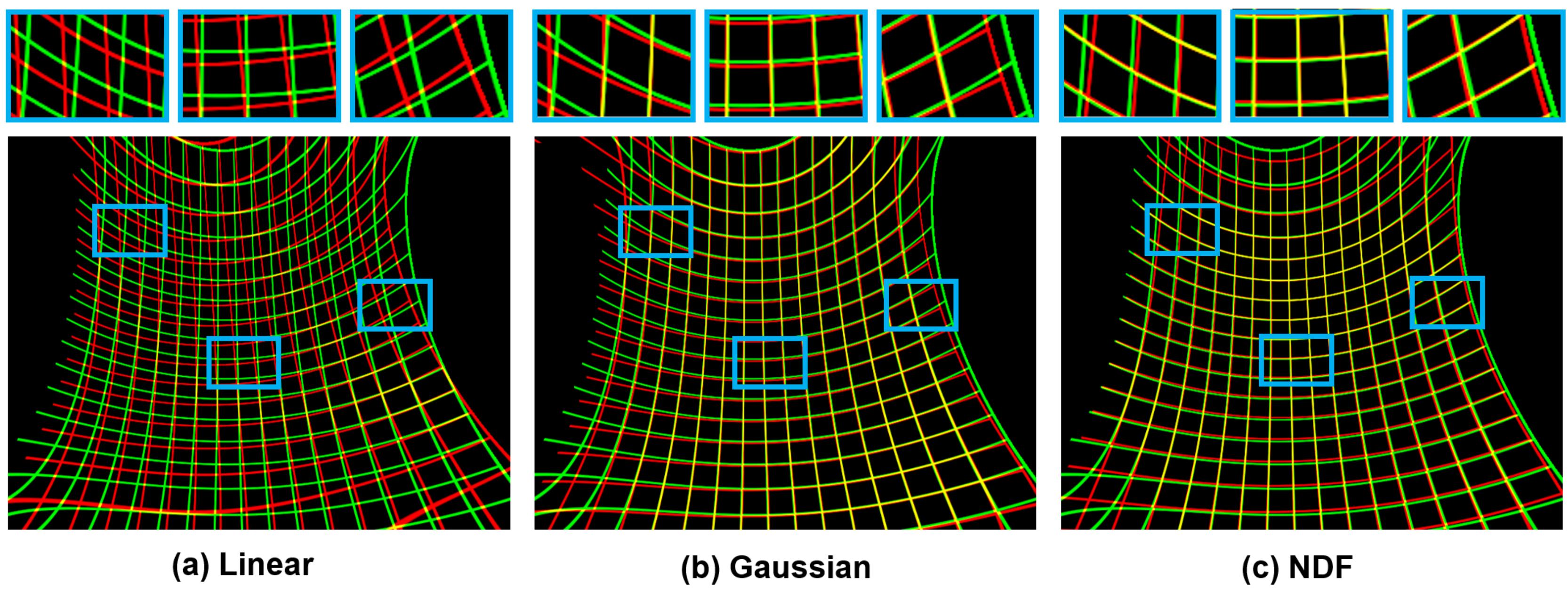}
 \caption{Qualitative comparison of estimated display coordinate from (a) linear interpolation, (b) Gaussian polynomial interpolation, and (c) NDF, on the viewpoint image near the center of eyebox. The green grids represent the ground truth, and the red grid is the display coordinates estimated from each method. The more the two coordinate systems matched, the closer the color of the grid approached yellowish. Three regions of interest (blue rectangles) are enlarged on each image.}
 \label{fig:result-vis-distortion}
\end{figure}

\begin{figure}[tb]
 \centering
 \includegraphics[width=\linewidth]{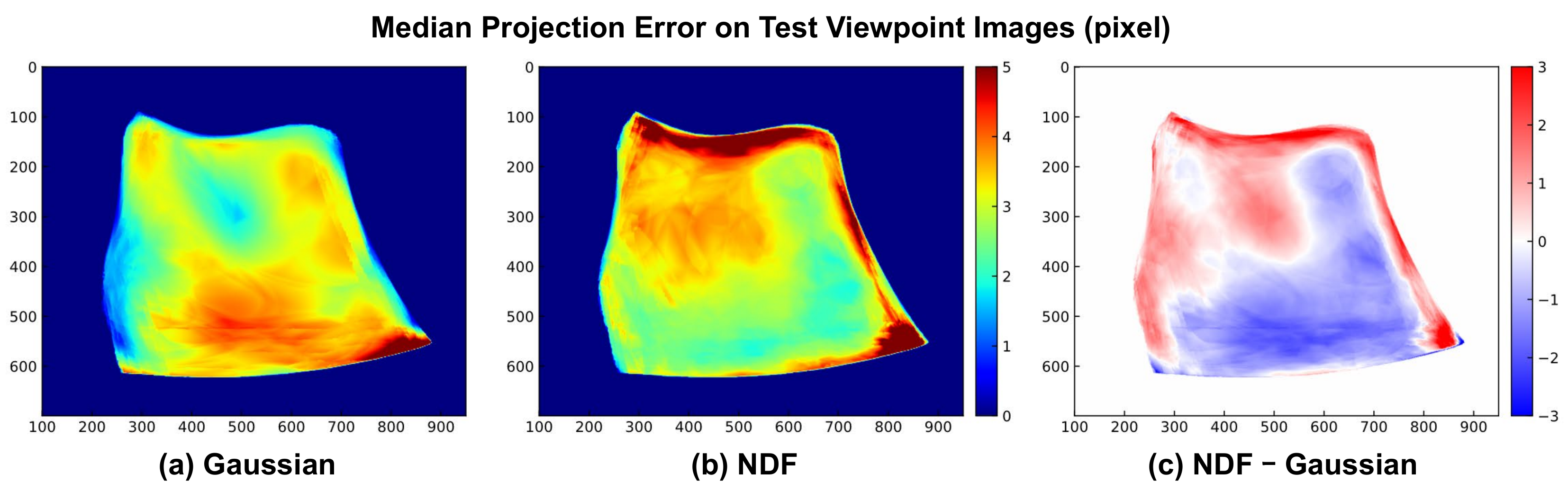}
 \caption{Error averaged over each pixel of the viewpoint camera. (a) Error from Gaussian fitting, (b) error from NDF, and (c) difference between the two. The red area indicates that Gaussian fitting is better, and the blue area indicates that NDF is better.}
 \label{fig:result-compare-images}
\end{figure}

We analyzed the properties of the distribution of reprojection errors perceived in the FoV by comparing (iii) Gaussian polynomial fitting and (iv) NDF.
Fig.~\ref{fig:result-vis-distortion} shows the difference between the ground truth and the actual transformation of the display coordinate system using the estimated distortion map.
From Fig.~\ref{fig:result-vis-distortion} (b), the estimated map with Gaussian fitting is accurate vertically but has relatively large horizontal deviations. In contrast, the estimated map from NDF (Fig.~\ref{fig:result-vis-distortion} (c)) shows a uniform fit in both horizontal and vertical directions near the center of the FoV, although the error is larger than Gaussian at the periphery of the FoV.

To evaluate the error distribution within the FoV, we calcuated pixel-wise average of the reprojection error in all test viewpoint images, as shown in Fig.~\ref{fig:result-compare-images}. While the Gaussian fitting does not have a smooth distribution of errors, NDF has smaller errors from the center to the lower right of the FoV, and the pixels with the largest errors are concentrated only at the periphery of the FoV. 
We also confirmed that the errors of NDF were better for 55~\% of the total FoV pixels.
This result shows that NDF learns the distortion of the target AR-NED well.

The error tends to be larger in the periphery of the FoV in NDF.
This is likely due to the fact that NDF learns not only the map but also the intensity of the map on a pixel-wise, \ie, the shape of the FoV.
As a result, NDF cannot cope with abrupt changes of the intensity $\probability$ at its periphery, resulting in large errors in the weighted sum (Eq.~(\ref{eq:volume-rendering})).
This problem could be addressed by combining NDF with some explicit model, \eg, explicitly defining the display surface in space in advance and sampling the surrounding area with NDF. Such a combination of NDF and explicit models to improve accuracy is further discussed in Sec.~\ref{sec:discussion}.

\subsection{Error Distribution depending on Viewpoint Position}\label{sec:experiments-difference-viewpoints}

\begin{figure}[tb]
 \centering
 \includegraphics[width=0.97\linewidth]{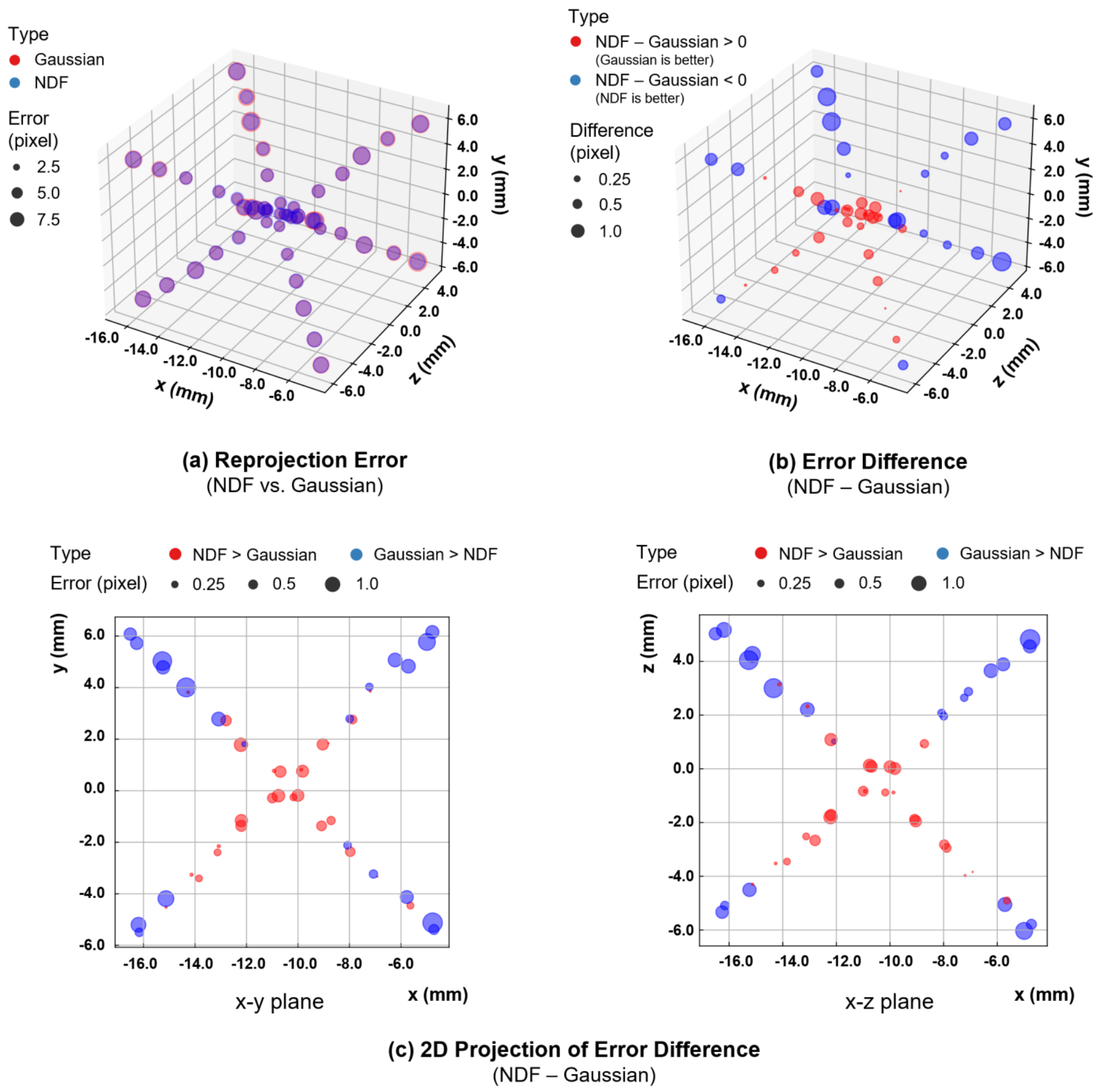}
 \caption{Comparison of reprojection errors against the distribution of 48 test viewpoints in the eyebox using 8 training viewpoints. (a) Median reprojection errors in the reconstructed distortion map at each viewpoint position. The color of each circle indicates the method (red: Gaussian, blue: NDF), and the radius indicates the error at each viewpoint position. (b) Difference of (a) reprojection errors between NDF and Gaussian at each test viewpoint. The red plot shows the viewpoint position at which Gaussian fitting reconstructs the distortion map more accurately, and the blue plot shows the opposite. (c) Scatter plots of the results in (b) projected onto the (Left) x-y plane and (Right) x-z plane.}
 \label{fig:result-error-distrib}
\end{figure}

Next, we evaluated the accuracy of the reconstruction of the distortion map with respect to changes in viewpoint $\EyePose$. 
With Gaussian fitting and NDF trained on 8 viewpoints, we calculate the reprojection error of each pixel test viewpoint position that is correspond to Fig.~\ref{fig:dataset-overview} (b).
Fig.~\ref{fig:result-error-distrib} (a) shows the median of the reprojection error on the entire pixels of the field of view at each test viewpoint position. To further clarify the difference between the two methods, Fig.~\ref{fig:result-error-distrib} (b) shows the distribution of the difference of the NDF reprojection error minus the Gaussian, as in Fig. ~\ref{fig:result-compare-images} (c). 
From Fig.~\ref{fig:result-error-distrib} (b), NDF shows better results at viewpoints far from the center of the eyebox. This means that Gaussian overfits the training data near the center of the eye box, while NDF is able to reproduce the distribution of the distortion map uniformly across the entire eye box. From the result, we confirmed that NDF can reproduce the distortion map robustly even when the viewpoint position changes.


\subsection{Error Analysis of Different Network Architecture}\label{sec:result-net-arch}
\begin{figure}[tb]
 \centering
 \includegraphics[width=\linewidth]{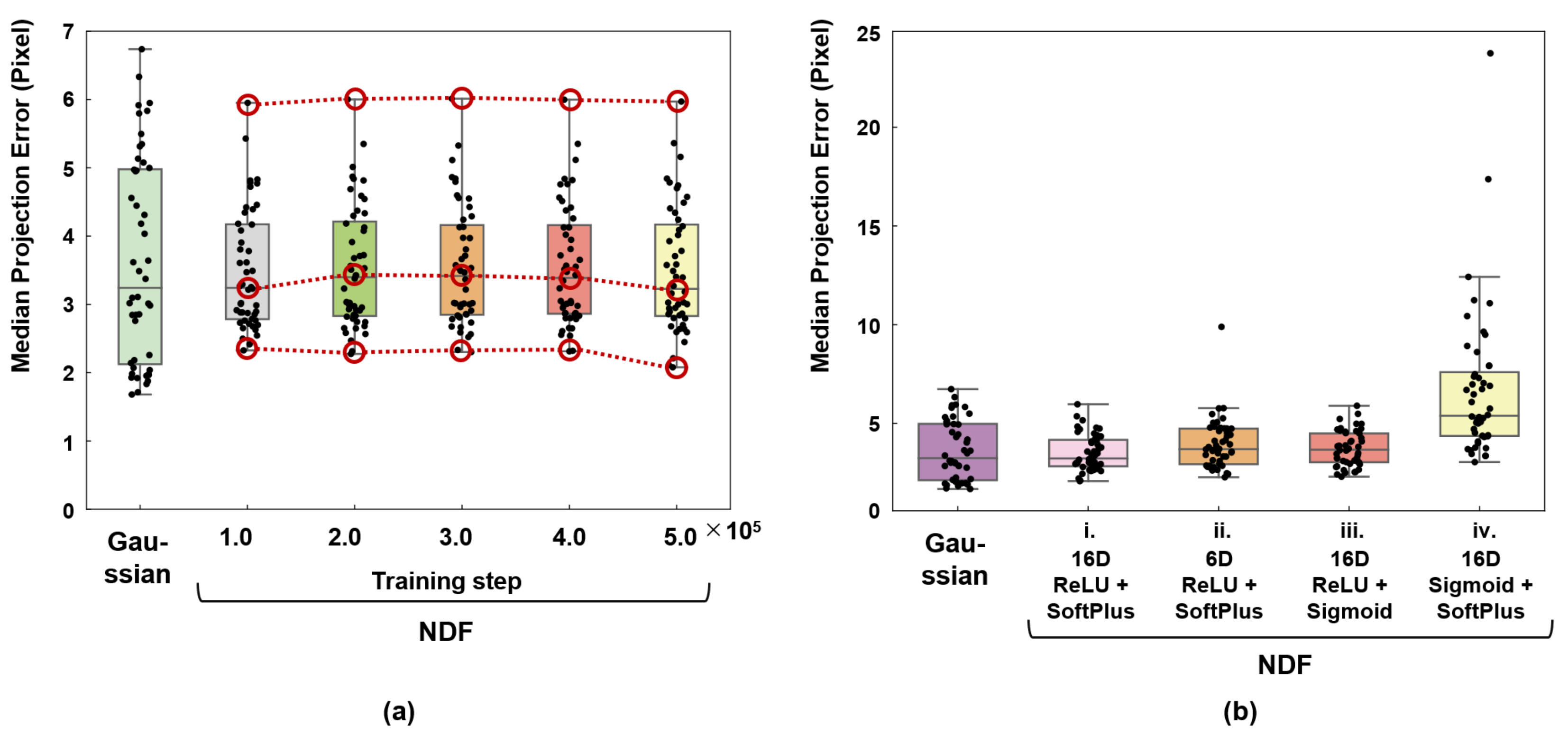}
 \caption{(a) The error of Gaussian interpolation and NDF with varying number of training steps from $1.0\times10^5$ to $5.0\times10^5$ using the 8 training viewpoints. (b) The errors of Gaussian interpolation for reference and NDFs for different combinations of positional encoding dimensions and output activation functions. Each label (i.) - (iv.) on the figure indicates, in order, the dimension of positional encoding, activation function for coordinates, and for intensity.}
 \label{fig:result-compare-training-step}
\end{figure}

Finally, we evaluate the effects of different network parameters on accuracy.
It is known that the number of layers and channels in the network has little effect on accuracy, while the number of dimensions of the position encoding, the activation function, and the number of training steps significantly impact accuracy~\cite{tancik2020fourfeat, sitzmann2020siren}. Thus, we trained NDFs under different conditions while varying these parameters and evaluated their accuracy on the training dataset ($N=8$).


\subsubsection{Training step.}
We compared networks trained with a different number of training steps from $1.0\times10^5$ to $5.0\times10^5$. During the experiment, the dimension of encoding $L=16$, the activation function of the coordinate MLP, and the intensity MLP were fixed to ReLU and SoftPlus, respectively.

Figure.~\ref{fig:result-compare-training-step} (a) shows the error of NDF at different numbers of training steps. When the number of steps is increased by $1.0\times10^5$, the mean error changes as follows: \{3.61, 3.78, 3.75, 3.72, 3.23\} pixel.
The results show that the minimum error value decreases as the number of training steps increases. However, there is no change in the intermediate error value for all viewpoints, confirming that the accuracy of NDF is as good as the Gaussian fitting even at $1.0\times10^{5}$ training steps. 
This result indicates that the NDF is already able to represent image distortion well in the early stages of training.
One possible reason for the result is that the image distortion reproduced by NDF is simpler in structure than that of NeRF, which targets natural images.

\subsubsection{Input dimension of Positional Encoding.}
We compared networks trained with a different encoding dimension ($L=6, 16$). We fixed number of learning steps at $5.0\times10^5$.

Figure~\ref{fig:result-compare-training-step} (b, i.) and (b, ii.) show the training results with the dimension of positional encoding $L$ set to 16 and 6. The mean errors of (b, i.) and (b, ii.) are 3.23 pixel and 3.62 pixel, respectively.
We previously assumed that reducing the number of encoding dimensions would not change the accuracy of NDF as discussed in Sec.~\ref{sec:positional-encoding}. However, from the result, as with the original NeRF, increasing the number of encoding dimensions reduced the error. This can be attributed to the fact that the current NDF learns not only the distortion of the image but also the range visible from the display in FoV, which results in higher frequencies at the periphery of the NED.

\subsubsection{Selection of Activation Functions.}
Finally, we trained the network with different combinations of activation functions for the coordinate MLP (ReLU or Sigmoid) for $\ImageCoords{\displayCoordinate}$ and the intensity MLP (SoftPlus or Sigmoid) for $\probability$. 

Figure.~\ref{fig:result-compare-training-step} (b, i.),  (b, iii.) and (b, iv.) show that the errors when varying the combination of output activation functions.
The mean errors of (b. i.). (b, iii.), and (b, iv) are 3.23 pixel, 3.60 pixel, and 5.36 pixel, respectively.
From the result, as expected in Sec.~\ref{sec:impl-network}, the accuracy was greatly improved by using ReLU for the output activation function of $\ImageCoords{\displayCoordinate}$. 
In constrast, changing the activation function of $\probability$ from Softplus to Sigmoid did not significantly improve the accuracy.
This result indicates that the virtual display surfaces of the wide-FoV NEDs targeted in this paper form multiple images through repeated multistage reflections and refractions.

\section{Limitation and Future Work}\label{sec:discussion}
From experiments, NDF can synthesize novel-view distortion maps for the wide-FoV AR-NED with accuracy equal to or better than explicit polynomial fitting models.
Our fully implicit, MLP-based approach is a completely different from existing works for distortion correction of NEDs. While the current NDF model is still rough around the edges, it has many possibilities for improvements and future research directions. This section discusses both such limitations and potential research directions.

\vspace{-1mm}
\paragraph{Real Time Dynamic Distortion Correction.}
As mentioned in Sec.~\ref{sec:intro}, dynamic distortion correction is one of the most important issues for NEDs, yet unsolved. In particular, real time distortion correction in response to eye tracking is required to make dynamic adjustments that are imperceptibly fast for the user.
Our NDF is compatible with real time distortion map generation. Thanks to neural network-based architecture, NDF can be GPU-accelerated. Moreover, since NED has almost the same configuration as NeRF, acceleration methods proposed in the research field of NeRF can be applied almost directly to NED.
For example, InstantNeRF~\cite{mueller2022instant} uses hash tables to adapt multi-resolution position encoding to GPU calculations, enabling the generation of $1920\times1080$ pixel resolution images at tens of milliseconds.
Since our initial desire in this paper is to verify our NDF concept and InstantNeRF is implemented on a customized CUDA kernel that is hard to customize, we currently have not implement NDF on its architecture. 
However, in theory, it is possible to run NDF on the real time framework.

\paragraph{Combination with Explicit Models.}
In this paper, we defined NDF as a fully implicit model that does not assume an a priori optical model.
However, since NDF is essentially a ray-casting-based method, it can be extended to a hybrid model, which combines NDF with conventional distortion correction methods that trace rays on explicitly defined optical models.
In the field of NeRF, some methods have been proposed to recover both the 3D shape and viewpoint-dependent texture of an object with high accuracy by intensively sampling points close to the object surface~\cite{oechsle2021unisurf, yariv2021volume}. 
In the same way, by intensively sampling points near the focal plane in a roughly defined optical design of NEDs, NDF can improve the distortion map's accuracy while fine-tuning the actual optical property of the NEDs.

\paragraph{Correction of Chromatic Aberration and Viewpoint-Dependent Blur.}
By extending the number of output dimensions, NDF could be utilized to calibrate various pixel-wise viewpoint-dependent properties, such as chromatic aberration~\cite{itoh2015semi} and viewpoint-dependent blur~\cite{itoh2016gaussian}. Although increasing the number of output dimensions may make learning difficult to converge, it can also include the correlation of each property, for example, implicitly learned, viewpoint-dependent color-mixing matrix for chromatic aberration.

\paragraph{Applying NDF tor Other Severely Distorted Optics.}
Although this paper only discusses the application of NDF to wide-FoV NEDs, it is expected that NDF can be applied to other non-smooth and extremely distorted optical systems, just as NeRF can be applied to images with abrupt changes in adjacent pixel values.
For example, NDF may be applied to aerial displays that form images using a special beam combiner~\cite{luo2017pepperscone}, or acquire correspondence between 3D scene and image coordinates in dynamic projection mapping.

\section{Conclusion}
We proposed NDF, an MLP-based distortion map generation method for wide-FoV NEDs. NDF implicitly learns virtual display surfaces as light source distributions in viewpoint-dependent space, a mutually complementary concept to explicit, geometric optics models.
Experiments show that NDF can synthesize distortion maps with an error of about 5.8 arcmins using only 8 training viewpoints, which is competitive with non-linear polynomial fittings. 
We also confirmed that NDF produces maps with better accuracy around the center of the FoV, and the accuracy improves as the number of training viewpoints increases.
NDF has the potential for higher accuracy by combination with explicit optical models, and real time distortion correction with GPU optimization.
We hope that our new approach will facilitate subsequent research and contribute to the realization of an immersive virtual experience that combines a wide field of view with perfect spatial consistency.

\begin{backmatter}
\bmsection{Funding} This project was partially supported by JST FOREST Grant Number JPMJFR206E, JST PRESTO Grant Number JPMJPR17J2, and JSPS KAKENHI Grant Number JP17H04692, JP20H05958, JP20H04222, and JP22J01340, Japan.
\bmsection{Acknowledgments} The authors thank Daisuke Iwai and Takumi Kaminokado for engaging in valuable discussions.
\bmsection{Disclosures} The authors declare no conflicts of interest.
\bmsection{Data availability} Data underlying the results presented in this paper are not publicly available at this time but may be obtained from the authors upon reasonable request.
\bmsection{Supplemental documents} This paper contains no supplemental document.
\end{backmatter}

\bibliography{main_revised.bib}






\end{document}